\pdfoutput=1

\documentclass[11pt]{article}

\usepackage{acl}

\usepackage{times}
\usepackage{latexsym}
\usepackage{kotex}

\usepackage[T1]{fontenc}

\usepackage[utf8]{inputenc}

\usepackage{microtype}

\usepackage{inconsolata}

\usepackage{graphicx}
\usepackage{algorithm}
\usepackage{algorithmic}
\usepackage{xcolor}
\usepackage{arydshln}
\usepackage{amssymb}
\usepackage{amsmath}
\usepackage{booktabs}
\usepackage{dsfont}
\usepackage{multirow}
\usepackage{tikz}
\usepackage{enumitem}
\usepackage{colortbl}
\title{\textsc{Mixture-of-Clustered-Experts}: Advancing Expert Specialization and Generalization in Instruction Tuning}

\author{Sugyeong Eo$^{1}$, Jungjun Lee$^{2}$, Chanjun Park$^{3*}$, Heuiseok Lim$^{1}$\thanks{Corresponding author}\\
$^{1}$Department of Computer Science and Engineering, Korea University, Republic of Korea\\
$^{2}$KT, Republic of Korea\\
$^{3}$School of Software, Soongsil University, Republic of Korea\\
\texttt{\{djtnrud,limhseok\}@korea.ac.kr \quad jungjun.lee@kt.com\quad chanjun.park@ssu.ac.kr}}

\begin{document}
\maketitle
\begin{abstract}
A sparse Mixture-of-Experts (MoE) architecture has emerged as a highly scalable solution by conditionally activating sub-modules without a proportional increase in computational costs. However, improving expert specialization to enhance performance and generalization remains a challenge for MoE, especially in instruction tuning scenarios characterized by significant input heterogeneity.
In this work, we propose the Mixture-of-Clustered-Experts (MoCE) to address this limitation through a dual-stage routing mechanism. The first stage in the mechanism performs expert group routing based on sequence-level features, while the second stage activates the top-$k$ experts within the group at the token level. This approach enables the effective partitioning of heterogeneous inputs based on their knowledge requirements, encouraging expert group specialization while maintaining the advantages of token-level routing.
We evaluate MoCE across a comprehensive set of benchmarks, demonstrating its consistent superiority over strong baselines and its enhanced generalization capabilities. Detailed analysis further highlights the robustness and effectiveness of MoCE.
\end{abstract}

\section{Introduction}
A sparse Mixture-of-Experts (MoE) has garnered significant attention for its ability to scale parameters with minimal computational overhead~\cite{shazeer2017,dai-etal-2024-deepseekmoe,jiang2024mixtral,xue2024openmoe}. MoE is a modified version of the Transformer architecture, replacing its feed-forward network (FFN) layer with an MoE layer that consists of multiple FFNs (experts). A gating function conditionally activates the most suitable experts for processing each token, enabling dynamic routing of input tokens among experts.
Harnessing the unique characteristics of MoE, its integration with instruction tuning has outperformed dense counterparts, establishing itself as a compelling approach in the field~\cite{komatsuzaki2023sparse,zadouri2024pushing,ostapenko2023case,zhu2024dynamic}.

However, instruction tuning in the MoE architecture still leaves room for improvement.
The input data for instruction tuning covers over a thousand NLP tasks and spans a wide range of domains~\cite{pmlr-v202-longpre23a,peng2023instruction,wang-etal-2022-super}.
The inherent heterogeneity of this input data poses challenges to developing expert specialization, where each expert acquires focused knowledge without overlapping with others~\cite{chen2023sparse,cai2024survey}. In the absence of such specialization, overlapping knowledge among experts leads to redundancy, and the complexity of input data forces individual experts to handle dispersed knowledge~\cite{dai-etal-2024-deepseekmoe}.

Furthermore, the routing mechanism in standard MoE models operates exclusively at the token level, where input features derived from sequence-level information, such as domain and task, indirectly influence the routing process. This restricts the model's ability to fully manage the complexity of input data~\cite{kudugunta-etal-2021-beyond-distillation,zhu2024dynamic}.
These limitations highlight the need to distinctly partition and process inputs effectively based on their knowledge requirements to encourage specialization while enhancing generalization performance. Since instruction tuning serves as a regularization technique to alleviate the overfitting issue in MoE fine-tuning~\cite{shen2024mixtureofexperts,dou2024loramoe}, the central challenge lies in enhancing the model's generalization capabilities while preserving these benefits.

In this work, we introduce the Mixture-of-Clustered-Experts (MoCE), an extension of the MoE architecture that incorporates a dual-stage routing mechanism leveraging both sequence-level and token-level information.
In MoCE, we modify the MoE structure by organizing experts into groups, with each group containing multiple experts. MoCE activates experts in two stages:
The first stage involves \textit{sequence-level expert group allocation}, where an expert group is selected based on cluster information derived from sequence embeddings. To achieve this, we employ a k-means clustering algorithm to partition inputs, leveraging the latent relationships among data points for more multifaceted segmentation. In the subsequent \textit{token-level expert allocation}, a group-specific gating function selectively activates a subset of experts from the activated expert group at the token level.
\textit{This approach enables specific expert groups to process specialized knowledge tailored to distinct input clusters while maintaining fine-grained expert activation optimized for handling individual tokens.}

Additionally, MoCE preserves the computational efficiency of MoE models by keeping the overall number of activated experts unchanged. However, MoCE still inherits a fundamental limitation of the MoE architecture: the VRAM-intensive requirement of loading all experts into memory. We adopt a lightweight adapter-based approach, as introduced by \citet{wu2024parameter}, making MoCE more practical for real-world applications.
We conduct extensive evaluations across diverse benchmarks, including mathematical problem-solving, code generation, reasoning, and knowledge-based tasks. The experimental results show that MoCE demonstrates superior performance not only over general baselines but also over models individually optimized for specific domains.
A comprehensive case study and analysis further validate the effectiveness and generalization capabilities of MoCE. Our contributions are threefold:
\begin{itemize}
    \item We propose the Mixture-of-Clustered-Experts (MoCE) to effectively partition inputs based on their knowledge requirements, inducing expert specialization.
    \item We ensure efficiency by preserving the number of activated experts and leveraging a fast-converging k-means clustering algorithm for sequence-level routing.
    \item Extensive experiments and analysis demonstrate the superior performance of MoCE, highlighting its practical potential for a wide range of real-world applications.
\end{itemize}

\section{Related Work}
The Sparse Mixture of Experts (MoE) architecture has become a foundational framework for scaling model parameters~\cite{shazeer2017,zoph2022st,ma2018modeling,jiang2024mixtral,raposo2024mixture}.
In particular, studies show that instruction tuning in MoE functions as a regularization technique that addresses the overfitting issue and outperforms dense counterparts, establishing it as a promising research direction~\cite{shen2024mixtureofexperts,komatsuzaki2023sparse,zadouri2024pushing,ostapenko2023case}.

With advancements in MoE, an increasing number of studies highlight the pivotal role of expert specialization~\cite{chen2023sparse,cai2024survey,shen2024mixtureofexperts}.
In the context of encouraging each expert to acquire distinct knowledge, studies have addressed the challenge of input heterogeneity by targeting specific criteria~\cite{dai2022one,zhong2022meta,ren2023pangu,sarkar2024revisiting,kudugunta-etal-2021-beyond-distillation}. For instance,~\citet{zhao2023sparse} introduce linguistic-guided routing with language-specific expert allocation, while \citet{gururangan-etal-2022-demix} sparsely activate domain-specific experts depending on the input domain.
\citet{dai2022one} define expert assignments for various input modalities.
Unlike these approaches that apply explicit routing criteria, our method leverages a clustering algorithm to capture latent input features and partition inputs accordingly.

Additionally, studies have explored routing mechanisms at the sequence or task level~\cite{gou2023mixture,kudugunta-etal-2021-beyond-distillation,fan2024towards}. While these studies focus on routing beyond the token level, our work employs a hierarchical routing strategy based on expert grouping. This enables the reflection of sequence-level information while preserving the benefits of selecting the most suitable experts to process individual tokens.

MoE has also been extended to parameter-efficient tuning techniques~\cite{wang-etal-2022-adamix,diao-etal-2023-mixture}, including the application of LoRA-based adaptations within the MoE framework~\cite{zadouri2024pushing,wu2024mixture}. Building on this direction, \citet{wu2024parameter} presents the adapter-based upcycled model adaptation method. Unlike the LoRA-based approach, this structure minimizes additional memory usage caused by weight merging and enables parallel computation, leading us to adopt it for improved efficiency.

\begin{figure*}
\centering
\includegraphics[width=\linewidth]{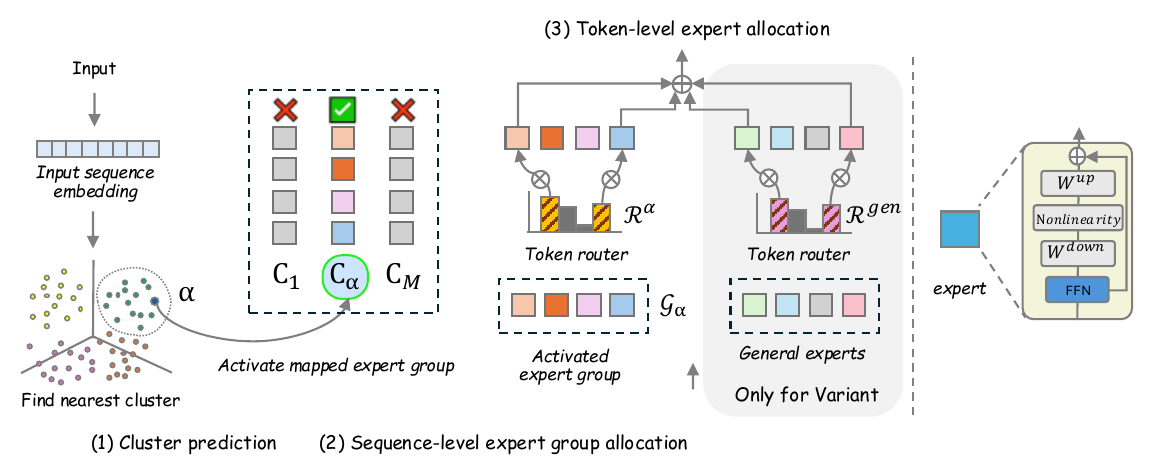}
\caption{The Overall Architecture of MoCE. MoCE consists of two hierarchical stages: (1) Sequence-level expert group allocation, where an input's sequence embedding is applied to predict a cluster, activating a corresponding expert group $\mathcal{G}^\alpha$ with its gating function $\mathcal{R}^\alpha$. (2) Token-level expert allocation, where $\mathcal{R}^\alpha$ computes routing probabilities for each expert on a per-token basis, selecting the top-$k$ experts to generate adapter outputs via weighted summation. An extended variant includes a general router $\mathcal{R}^{gen}$ and general experts, whose outputs are fused with the adapter outputs to form the final representation.} \label{fig:main}     
\end{figure*}

\section{\textsc{Mixture-of-Clustered-Experts}}
We propose the Mixture-of-Clustered-Experts (MoCE) architecture, which employs a hierarchical dual-stage routing mechanism that operates at both the sequence and token levels. In this section, we provide a detailed description of the MoCE.

\subsection{Preliminaries}
The MoE architecture replaces the dense feed-forward network (FFN) sub-layers within the Transformer block with MoE layers. Each MoE layer consists of a set of FFNs, denoted as $\{\mathcal{E}_i\}_{i=1}^N$. 
A gating function, also referred to as a router $\mathcal{R}(\cdot)$, sparsely activates the experts by routing the intermediate token representation $x$ in the input sequence $s$ to the most appropriate experts.

For the token representation, which is the output of the multi-head attention sub-layer, the router logit $h(x) = W_G^\top \cdot x$ is computed through a linear projection, where $W_G \in \mathbb{R}^{d_{model} \times N}$ denotes a trainable projection matrix. These scores are normalized via a softmax function over the $N$ experts:
\begin{equation}
    \mathcal{R}(x)_i=\frac{\exp{(h(x)_i)}}{\sum_{j=1}^N \exp{(h(x)_j)}},
\end{equation}
where the output of the gating function serves as the routing weight. The final output $y$ is calculated as a weighted combination of the routing weights and the outputs of the activated experts:
\begin{equation}\label{eq:weight_sum}
y=\sum_{i=1}^N\text{TopK}(\mathcal{R}(x)_i,k) \cdot \mathcal{E}_i(x),
\end{equation}
where the TopK function determines top-$k$ experts to route the token $x$:
\begin{equation}\label{eq:selecttopk}
\text{TopK}(\mathcal{R}(x)_i, k) = \begin{cases}\mathcal{R}(x)_i & \text{if } i \text{ is in the } \\ 
\hspace{0em} & \text{top-$k$ of } \mathcal{R}(x), \\
0 & \text{otherwise.}\end{cases}
\end{equation}

To address the high VRAM demands of MoE, we leverage the Parameter-Efficient Sparsity Crafting (PESC) proposed by~\citet{wu2024parameter}. We initialize model parameters from a pre-trained dense model~\cite{komatsuzaki2023sparse} and integrate adapters into each FFN. Therefore, the outputs in Equation~(\ref{eq:weight_sum}) are modified as $y=\sum_{i=1}^NTopK(\mathcal{R}(x)_i, k)\cdot \mathcal{A}_i(x)$,
where we denote each adapter $\mathcal{A}_i(x)=\sigma(W_i^{\text{down}}\cdot \mathcal{E}(x))\cdot W_i^{\text{up}}+x$ as expert.

\subsection{Expert Grouping and Clustering}
\paragraph{Expert Grouping}
The MoCE layer is characterized by its grouping of experts, where multiple experts are combined to form an expert group. Each expert group $\mathcal{G}_j$ $(j=1,...,M)$ comprises $N$ experts, denoted as $\{\mathcal{A}_i^{\mathcal{G}_j}\}_{i=1}^N$, alongside a group-specific gating function $\mathcal{R}^j$.

\paragraph{Clustering}
We opt for a k-means clustering mechanism to partition the input according to the features derived from the sequence-level. This allows for the input processing based on multidimensional attributes rather than relying solely on observable features.
To enable cluster predictions for inputs during training and inference, we train a clustering model in advance.

Let $\mathcal{C}$ represent the set of clusters and $e \in E_\alpha$ denotes a sequence embedding assigned to cluster $\alpha$. We utilize an encoder-based embedding model~\cite{wang2022text} to generate these embeddings, capturing the contextual representation of the entire sequence. The clustering objective $J$ minimizes the $L^2$ distance between each embedding $e$ and its corresponding centroid $\mu_\alpha$, formulated as $ J=\sum_{c=1}^\mathcal{|C|}\sum_{e\in E_\alpha}||e - \mu_\alpha||^2$.
Each sequence embedding is assigned to the nearest cluster, and centroids are updated by averaging the embeddings in each cluster, as follows:
\begin{equation}
E_\alpha=\{e:||e-\mu_\alpha||^2 \leq ||e-\mu_c||^2, \forall c = 1,...,|\mathcal{C}|\},
\nonumber \end{equation}
\begin{equation}
\mu_\alpha = \frac{1}{|E_\alpha|}\sum_{e\in E_\alpha}e.   
\end{equation}
This iterative process continues until cluster assignments converge. 
Instead of manually setting the cluster count, we use the elbow method to determine the optimal number of clusters. We incrementally increase the cluster count and identify the point where the reduction in the sum of $L^2$ distances distinctly decelerates. Details about the elbow method are provided in the Appendix~\ref{app:elbow_method}.
While k-means clustering is often regarded as outdated, we prioritize its rapid convergence without hindering the efficiency of the MoE. 
Its linear complexity relative to the number of data points ensures scalability to larger datasets.

\subsection{Routing Mechanism}
Building upon the grouped expert structure and clustering model, MoCE introduces a hierarchical dual-stage routing mechanism: \textit{sequence-level expert group allocation} followed by \textit{token-level expert allocation}. Figure~\ref{fig:main} provides a detailed overview of the MoCE architecture.

\paragraph{Sequence-Level Expert Group Allocation}
We begin by performing sequence embeddings on the input samples. As noted earlier, we employ an encoder-based embedding model, which operates independently of the MoE's input token representation. This approach addresses the limitations of decoder-based models that rely solely on preceding tokens~\cite{behnamghader2024llmvec}. Each sequence embedding is then assigned a cluster number through a k-means clustering model, which identifies the nearest centroid relative to the input.

We then map the predicted cluster to a corresponding expert group. For example, if the assigned cluster number for input is $\alpha$, we map this to the expert group $\mathcal{G}_\alpha$. This process involves activating different expert groups based on the distinct characteristics of each cluster, with a one-to-one mapping, i.e., $|\mathcal{C}|=|\mathcal{G}|$.
Accordingly, the input is initially routed to the designated expert group corresponding to the assigned cluster number. 
Note that all token representations in the sequence are routed to their corresponding expert group, and only the group is activated to encourage specialization.
\begin{table*}[]
\resizebox{\textwidth}{!}{%
\begin{tabular}{lcccccc}
\toprule[1.5pt]
\multicolumn{1}{c}{\multirow{2}{*}{\textbf{Method}}} & \multicolumn{4}{c}{\textbf{Code}} & \multicolumn{2}{c}{\textbf{Mathematics}} \\ \cline{2-7}
& \textbf{HumanEval(@1)} & \textbf{HumanEval(@10)} & \textbf{MBPP(@1)} & \textbf{MBPP(@10)} & \textbf{MathQA} & \textbf{GSM8K} \\ \midrule[1.5pt]
Vanilla LLaMA & 13.57 & 17.72 & 15.57 & 19.67 & 28.81 & 23.35 \\
LLaMA (A) & 14.90 & 18.52 & 16.50 & 19.35 & 28.91 & 21.91 \\
BTX (A) & 13.00 & 17.52 & 17.53 & 20.24 & 28.71 & 22.67 \\
PESC & 16.00 & 23.21 & 20.97 & 25.77 & 29.25 & 33.21\\ \cdashline{1-7}
Specialized-Math & 14.28 & 19.46 & 15.28 & 21.83  & 28.17 & \underline{40.11}\\
Specialized-Code & 18.39 & 24.57 & 22.91 & 27.15 & 29.35 & 22.37 \\
Specialized-R\&K & 15.19 & 20.87 & 17.73 & 22.73 & 29.28 & 26.54 \\ \cdashline{1-7}
MoCE-E5 & 19.28 & \underline{30.05} & 22.11 & \underline{27.69} & \underline{30.35} & \textbf{41.93} \\  
MoCE-instructor & \underline{20.08} & 28.48 & \textbf{23.55} & \textbf{28.11} & 30.12 & 36.69\\
 MoCE-instructor(V)&	\textbf{21.75} & \textbf{31.90}& \underline{23.12}& 27.21 & \textbf{30.75} & 37.15 \\\midrule[1.5pt]
\multicolumn{1}{c}{\multirow{2}{*}{\textbf{Method}}} & \multicolumn{2}{c}{\textbf{General}} & \multicolumn{3}{c}{\textbf{Knowledge \& Reasoning}} & \multicolumn{1}{||c}{\cellcolor[HTML]{DADCFD}}
\\ \cline{2-6}
& \textbf{BBH} & \textbf{MMLU-Pro} & \textbf{Winogrande} & \textbf{ARC(Easy)} & \textbf{ARC(Challenge)} & \multicolumn{1}{||c}{\multirow{-2}{*}{\cellcolor[HTML]{DADCFD}\textbf{Average}}} \\ \midrule[1.5pt]
Vanilla LLaMA & 36.76 & 20.02 & 66.46 & 69.70 & 44.28 & \multicolumn{1}{||c}{\cellcolor[HTML]{DADCFD}32.36} \\
LLaMA (A) & 36.96 & 19.03 & 66.38 & 68.90 & 43.52 & \multicolumn{1}{||c}{\cellcolor[HTML]{DADCFD}32.26} \\
BTX (A) & 37.13 & 20.06 & 66.22 & 69.61 & 44.20 & \multicolumn{1}{||c}{\cellcolor[HTML]{DADCFD}32.44} \\
PESC & 37.94 & 20.81 & 67.56 & 71.59 & 45.56 & \multicolumn{1}{||c}{\cellcolor[HTML]{DADCFD}35.62} \\ \cdashline{1-7}
Specialized-Math & 38.29 & 19.27 & 67.80 & 69.49 & 43.26 & \multicolumn{1}{||c}{\cellcolor[HTML]{DADCFD}34.29} \\
Specialized-Code & 37.58 & 19.55 & 67.01 & 70.20 & 44.97 & \multicolumn{1}{||c}{\cellcolor[HTML]{DADCFD}34.91} \\
Specialized-R\&K & 38.80 & \underline{21.00} & \underline{68.03} & 70.96 & \textbf{46.84} & \multicolumn{1}{||c}{\cellcolor[HTML]{DADCFD}34.34} \\ \cdashline{1-7}
MoCE-E5 & \textbf{39.56} & 20.21 & \textbf{68.59} & \underline{71.80} & \underline{46.33} & \multicolumn{1}{||c}{\cellcolor[HTML]{DADCFD}\textbf{37.99}} \\ 
MoCE-instructor & \underline{39.09} & \textbf{21.10} & 67.80 & \textbf{72.26} & 46.25 & \multicolumn{1}{||c}{\cellcolor[HTML]{DADCFD}37.61} \\ 
 MoCE-instructor (V) & 38.66 & 20.22 & \underline{68.43} & 71.00 & \underline{46.33} & \multicolumn{1}{||c}{\cellcolor[HTML]{DADCFD}\underline{37.87}} \\ \bottomrule[1.5pt]
\end{tabular}}\caption{Performance of MoCE across four evaluation categories\label{tb:main}}
\end{table*}

\paragraph{Token-Level Expert Allocation}
The second stage involves routing individual token representation to specific experts, maintaining the advantage of MoE in selecting experts best suited for processing the current token.
This token-level routing follows an equivalent approach to that of standard MoE models. One distinction is that this token-level routing is applied exclusively to the activated expert group. The outputs from the selected experts are computed as follows:
\begin{equation}
    y=\sum_{i=1}^N \text{TopK}(\mathcal{R}^{\alpha}(x)_i,k) \cdot \mathcal{A}^{\mathcal{G}_\alpha}_i(x).
\end{equation}
Selecting top-$k$ experts is mainly based on $k=2$, with soft merging additionally investigated as a complementary technique.~\cite{zadouri2024pushing,muqeeth2024soft}. In the case of soft merging, the outputs from each adapter are aggregated through a weighted combination, with the weights generated by the group-specific gating function $\mathcal{R}^\alpha$~\cite{puigcerver2024from}. This formulation enables exact gradient computation based on estimated gradients.

\subsection{MoCE Variant}
We introduce a variant of MoCE that includes additional general experts, $\{\mathcal{A}^{gen}_j\}_{j=1}^N$, and a general router, $\mathcal{R}^{gen}$. These processes all sequences, similar to the standard MoE model. As shown in Figure~\ref{fig:main}, this operates alongside the sentence-level and token-level routing, aiming to integrate knowledge obtained from all inputs. The output from the general experts is combined with the output from experts activated by the group-specific gating function, which is computed as:
\begin{align}
    y &= \sum_{i=1}^N \text{TopK}\left( \mathcal{R}^{\alpha}(x)_i, k \right) \cdot \mathcal{A}^{\mathcal{G}_\alpha}_i(x) \nonumber \\
    &\quad + \sum_{j=1}^N \text{TopK}\left( \mathcal{R}^{gen}(x)_j, k \right) \cdot \mathcal{A}^{gen}_j(x).
\end{align}
By combining both outputs, we merge the knowledge acquired from the entire dataset with specialized knowledge.

\section{Experiments}
\subsection{Experimental Setup}

\paragraph{Datasets.} 
We conduct instruction tuning on MoCE using three distinct datasets. These include SlimOrca~\cite{lian2023slimorca}, a curated version of the OpenOrca dataset comprising 518K multi-task examples; Magicoder~\cite{wei2024magicoder}, which contains 110K coding problems, providing a rich source for programming and algorithmic tasks; and MetaMathQA~\cite{zhong2022meta}, consisting of 395K mathematical questions. In total, approximately 1M data points are used to train the models in our experiments.

\paragraph{Models.}
To mitigate the limitations of decoder-only models, we opt for Instructor~\cite{su-etal-2023-one} and E5~\cite{wang2022text}, an encoder-based sequence embedding model. By applying the elbow method, we determine the optimal number of clusters and corresponding expert groups to be four and seven, with each group comprising four experts. For the backbone model, we train the LLaMA2~\cite{touvron2023llama} model with 7 billion parameters, selected for its wide applicability. To prevent expert collapse, we apply a load-balancing loss throughout the experiments. Further implementation details are provided in Appendix~\ref{appendix:implementation_details}.

\begin{figure*}
\centering
\includegraphics[width=\linewidth]{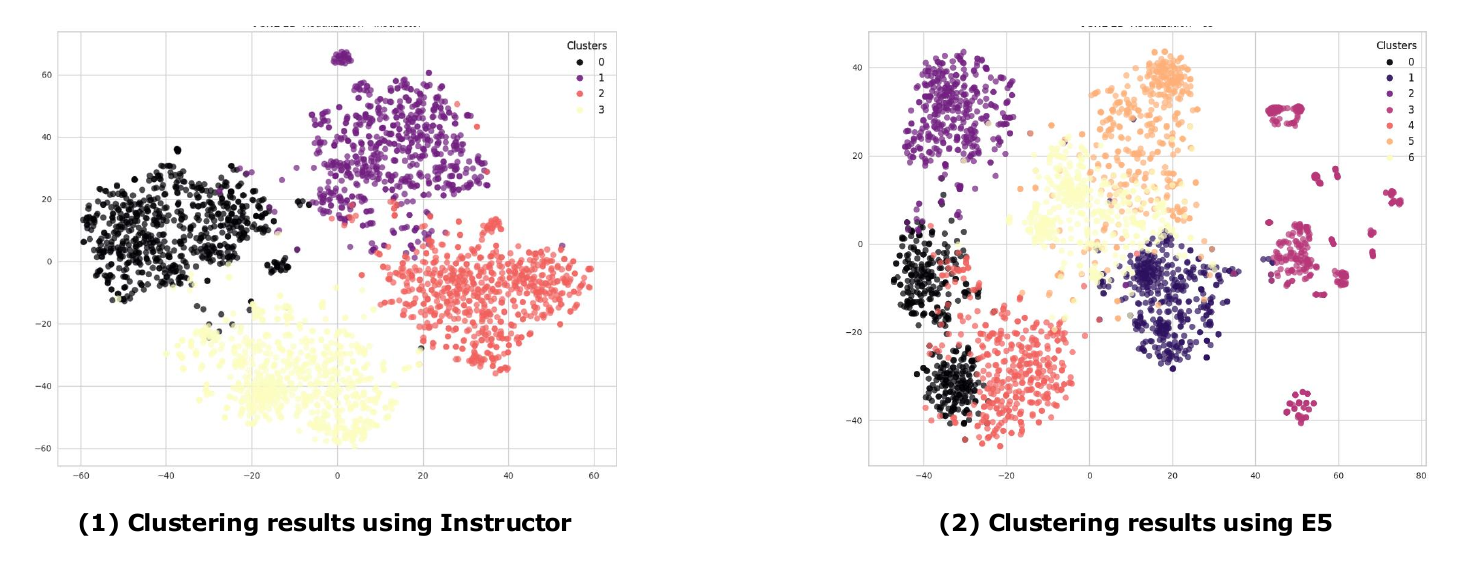}
 \caption{K-means clustering results based on sequence embeddings from Instructor and E5 models} \label{fig:embed} 
\end{figure*}

\begin{figure}
\centering
\includegraphics[width=\linewidth]{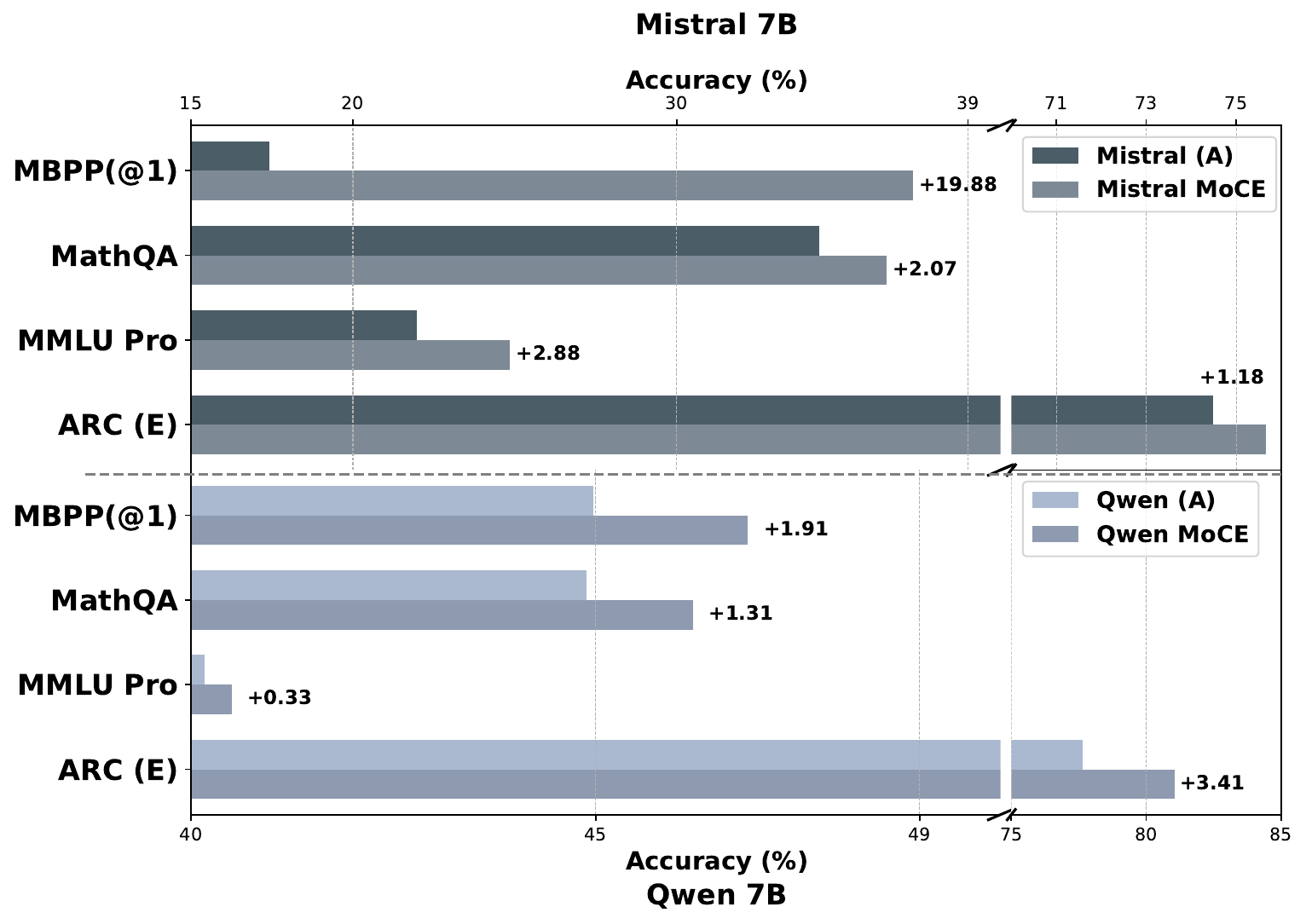}
 \caption{Performance of MoCE applied to Mistral and Qwen language models} \label{fig:mistral_barplot}
\end{figure}

\paragraph{Baselines.}
Given that MoCE adopts adapter-based training, we compare it against the following baselines: (1) LLaMA-Adapter, denoted as LLaMA (A), an adapter-based model designed for efficient parameter adaptation~\cite{houlsby2019parameter}, and (2) BTX~\cite{sukhbaatar2024branchtrainmix}, an MoE architecture that integrates the FFNs of independently trained domain-specialized LLMs, and subsequently learns a gating function to route inputs to the appropriate experts. To align with our experimental setting, we utilize its adapter-based implementation, denoted as BTX (A). (4) Parameter Efficient Sparsity Crafting (PESC)~\cite{wu2024parameter}, which integrates MoE architecture with adapters to enhance computational efficiency. To further evaluate MoCE's ability to balance specialization and generalization, we include domain-specific PESC models as additional baselines, each trained exclusively on tasks from mathematics, code, and reasoning \& knowledge domains.

\paragraph{Evaluation Benchmarks.}
To rigorously assess the effectiveness of MoCE, we employ a comprehensive benchmark spanning four distinct task categories. \textbf{Mathematics:} This category evaluates the model's ability to solve multi-step arithmetic and algebraic problems, using MathQA~\cite{amini-etal-2019-mathqa} and GSM8K~\cite{cobbe2021training}. \textbf{Code:} Coding proficiency and algorithmic reasoning are assessed through HumanEval~\cite{chen2021evaluating} and MBPP~\cite{austin2021program}. \textbf{Knowledge:} General knowledge and domain-specific understanding are evaluated leveraging BBH~\cite{suzgun2022challengingbigbenchtaskschainofthought} and MMLU-Pro~\cite{wang2024mmluprorobustchallengingmultitask}, which spans a wide range of subjects. \textbf{Reasoning:} The model's capacity for commonsense and logical reasoning is measured using Winogrande~\cite{10.1145/3474381} and ARC (easy and challenge subsets)~\cite{clark2018think}.
All evaluations are performed under the same settings using lm-evaluation-harness~\cite{eval-harness} and bigcode-evaluation-harness~\cite{bigcode-evaluation-harness}. The few-shot examples are sourced from lm-evaluation-harness.

\section{Results and Discussion}
\subsection{Main Results}
We compare MoCE against eight benchmark datasets spanning four task categories. Figure~\ref{fig:embed} presents the embedding results, which we detail in Appendix~\ref{app:embed}.
As shown in Table~\ref{tb:main}, MoCE consistently outperforms all baseline models on average. MoCE achieves an average score of 37.99 with MoCE-E5 and 37.61 with MoCE-Instructor, outperforming the baseline models LLaMA (A), which scores 32.26, BTX (A) at 32.44, and PESC at 35.62. MoCE exhibits particularly strong performance in the mathematics and code domains, while also maintaining high accuracy across the majority of evaluated benchmarks. Specifically, MoCE-E5 demonstrates superior performance over PESC, with scores improving from 33.21 to 41.93 on GSM8K and from 16.00 to 19.28 on HumanEval.

Compared to domain-specialized models trained on a single task category, MoCE demonstrates a clear advantage.  Although specialized models achieve strong performance within their respective domains, the performance diminishes on out-of-domain tasks. In contrast, MoCE not only matches or surpasses in-domain performance but also generalizes effectively across diverse task categories. These findings reveal that our dual-stage routing mechanism enables expert specialization by successfully managing heterogeneous inputs while preserving the regularization benefits of instruction tuning. Furthermore, the variant of MoCE demonstrates higher average performance compared to the standard MoCE. This indicates that the inclusion of general experts in MoCE improves performance by enabling effective collaboration with specialized experts.

\begin{table}[]
\centering
\resizebox{.5\textwidth}{!}{%
\begin{tabular}{ccccc|c}
\toprule[1.5pt]
\textbf{Model} & \textbf{MBPP(@1)} & \textbf{GSM8K} & \textbf{BBH} & \textbf{ARC(E)} & \textbf{Avg}\\ \midrule[1.5pt]
Qwen7B-Chat & 37.40 & 54.12 & 46.05 & 63.43 & 50.25\\ 
DeepSeek7B-Chat & 39.00 & 16.60 & 34.87 & 70.79 & 40.32\\ 
Mistral7B-Inst & 36.09 & 42.76 & 45.88 & 73.40 & 49.53\\ \hline
QwenMoE  & 36.60 &  51.71  & 41.69 & 68.31 & 49.58 \\ 
DeepSeekMoE  &	39.20 & 16.91 &	33.80 	& 73.15 & 40.77\\
\rowcolor[HTML]{DADCFD}Mistral-MoCE &	\textbf{41.05} & \textbf{54.97} 	&\textbf{48.43} &\textbf{75.67} & \textbf{55.03}\\ \bottomrule[1.5pt]
\end{tabular}%
}
\caption{Comparative evaluation results with dense and MoE models}
\label{tb:comparative_moe}
\end{table}
\subsection{Application of MoCE Across Different Model Families}
As illustrated in Figure~\ref{fig:mistral_barplot}, MoCE consistently outperforms the baseline across both the Mistral and Qwen model families. Notably, it achieves substantial improvements on domain-specific benchmarks, while also delivering consistent gains in general and reasoning tasks. On average, MoCE improves accuracy by +6.50 points in Mistral 7B and by +1.74 points in Qwen 7B.

These results indicate that MoCE enhances performance across diverse domains without exhibiting signs of overfitting to any particular training domain. We attribute these gains to MoCE’s structured routing mechanism, which facilitates targeted expert activation and promotes effective expert specialization, enabling robust adaptation to heterogeneous task distributions. These findings indicate that MoCE serves as a robust and transferable solution across model families beyond LLaMA.

\subsection{Evaluating MoCE Against Dense and MoE Models}
To assess the practical utility of MoCE, we conduct a comparative analysis with publicly released dense and MoE-based models.
Table~\ref{tb:comparative_moe} provides a comparative evaluation of the Mistral-MoCE model with respect to two categories of LLMs: dense models of similar size and MoE-based models. Compared to the dense Mistral 7B model, MoCE achieves substantial performance improvements, increasing GSM8K accuracy from 42.76 to 54.97 and MBPP from 36.09 to 41.05, along with consistent gains on other evaluation benchmarks.
Mistral-MoCE achieves consistently superior performance over both Qwen1.5-MoE and DeepSeekMoE, indicating a clear performance margin. These findings suggest that MoCE offers an effective solution for expert specialization and generalization, while retaining computational efficiency.

\begin{figure}
\centering
\includegraphics[width=\linewidth]{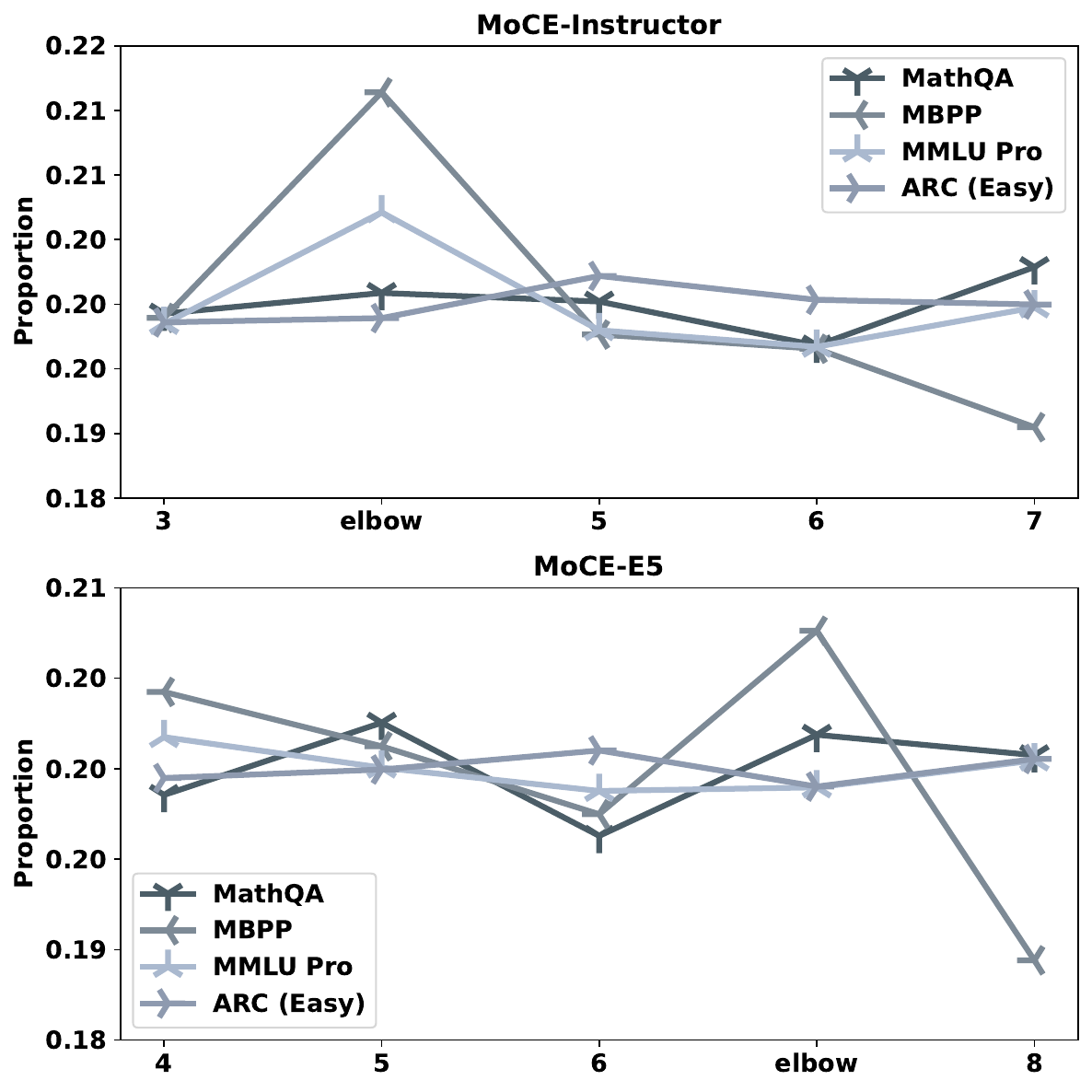}
 \caption{Performance variations across different numbers of expert groups, which correspond to the number of clusters} \label{fig:num_cluster}
\end{figure}

\begin{figure*}
\centering
\includegraphics[width=\linewidth]{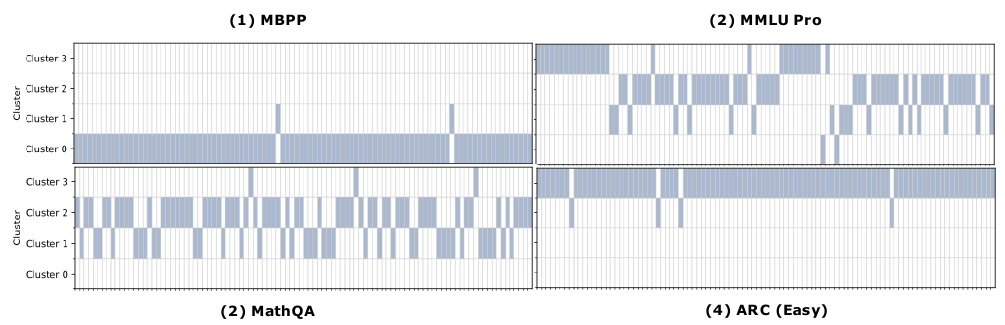}
 \caption{Cluster prediction results on four evaluation benchmarks, based on 100 sampled instances per benchmark. Sequence embeddings are obtained using the Instructor model, and k-means clustering is performed with four clusters.} \label{fig:cluster_pred}    
\end{figure*}

\begin{table}[]
\resizebox{\linewidth}{!}{%
\begin{tabular}{lcccc|c}
\toprule[1.5pt]
\multicolumn{1}{c}{\textbf{\# Experts}} & \multicolumn{1}{c}{\textbf{MBPP(@1)}} & \multicolumn{1}{c}{\textbf{MathQA}} & \multicolumn{1}{c}{\textbf{MMLU-Pro}} & \multicolumn{1}{c}{\textbf{ARC-Easy}} & \multicolumn{1}{|c}{\textbf{Average}} \\ \midrule[1.5pt]
4 experts (4*1) & 20.62 & 29.65 & 20.05 & 71.89 & 35.55 \\
8 experts (4*2) & 21.90 & 29.61 & 20.46 & 71.63 & 35.90 \\
\rowcolor[HTML]{DADCFD}16 experts (4*4) & \textbf{23.55} & \textbf{30.12} & \textbf{21.10} & \textbf{72.26} & \textbf{36.76} \\ \bottomrule[1.5pt]
\end{tabular}%
}
\caption{Comparison of configurations with 1, 2, and 4 experts per cluster in a four-cluster model}
\label{tb:expert}
\end{table}

\subsection{Discussion}
\paragraph{The elbow method successfully identifies the optimal number of expert groups.}
We apply the elbow method to determine the optimal number of clusters for MoCE and further analyze performance variations under different cluster configurations. Figure~\ref{fig:num_cluster} illustrates the performance trends as the number of clusters changes. Note that the number of clusters is equal to the number of expert groups. The elbow method identifies four and seven as the optimal cluster counts for Instructor and E5 embeddings, respectively, both yielding the best overall performance. These findings validate the elbow method as a reliable approach for selecting the optimal number of expert groups.

\paragraph{The number of experts within each expert group exhibits scalability.}
One of the key advantages of the MoE architecture is its scalability~\cite{zadouri2024pushing}. To evaluate this property, we conduct experiments by progressively increasing the number of experts within each expert group while keeping the number of clusters fixed. Table~\ref{tb:expert} presents the results with expert counts ranging from 4 (four clusters with one expert each) to 16 (four clusters with four experts each) per cluster. As shown in the table, increasing the number of experts consistently improves model performance. In particular, the average score increases steadily from 35.55 to 36.76 as the expert count scales up, demonstrating that MoCE effectively leverages additional experts to achieve higher performance.

\paragraph{Assigning inputs to distinct expert groups is essential for achieving effective expert specialization.}

Figure~\ref{fig:cluster_pred} presents the distribution of cluster predictions across the four domains.
The results reveal distinct cluster preferences depending on the domain. For example, MBPP inputs are predominantly assigned to cluster 0, while ARC samples are mostly associated with cluster 3. Mathematics examples are largely concentrated in clusters 2 and 3. In the case of MMLU, the cluster predictions are more evenly distributed, which aligns with the dataset’s diverse coverage of world knowledge.
These findings suggest that expert specialization in MoCE successfully emerges from assigning inputs to specific expert groups based on their sequence-level features. 

\begin{table}[]
\resizebox{.5\textwidth}{!}{%
\begin{tabular}{lcccc}
\toprule[1.5pt]
\multicolumn{1}{c}{\textbf{Method}} & \textbf{MBPP(@1)} & \textbf{MathQA} & \textbf{MMLU-Pro} & \textbf{ARC(E)}  \\ \midrule[1.5pt]
\multicolumn{5}{c}{\textit{routing strategy}}\\\hline
\rowcolor[HTML]{DADCFD} w/ Top-2 & \textbf{23.55} & 30.12 & \textbf{21.10} & 72.26 \\
w/ Top-1 & 22.33 & 29.75 & 20.34 & 70.96 \\
w/ Soft Merging & 22.93  & \textbf{30.39}& 20.64 & \textbf{72.31} \\ \hline
\multicolumn{5}{c}{\textit{top-k token routing strategy}}\\\hline
\rowcolor[HTML]{DADCFD} MoCE & \textbf{23.55} & \textbf{30.12} & \textbf{21.10} & \textbf{72.26} \\
w/o Clustering & 21.72  & 29.01& 20.68 & 71.13 \\
\multicolumn{1}{c}{w/o Token routing} & 20.62 & 29.65 & 20.05 & 71.63 \\\bottomrule[1.5pt]
\end{tabular}%
}
\caption{Ablation study on routing strategies in MoCE}
\label{tb:ablation}
\end{table}

\paragraph{Dual-stage routing facilitates effective expert specialization in MoCE.}
Each component of the MoCE framework plays an important role in enabling effective expert specialization. We validate this via an ablation study shown in Table~\ref{tb:ablation}.
We begin by analyzing the impact of varying the value of $k$ in the Top-$k$ token routing mechanism. Among the configurations evaluated, Top-2 routing and soft merging yield higher performance. These results suggest that selecting multiple experts per token, rather than relying solely on the most confident single expert, contributes to more effective expert utilization in MoCE. 

To gain a deeper understanding of the effectiveness of the hierarchical dual-stage routing structure, we selectively disable each stage. By holding all other variables constant, we isolate the effect of each architectural component and observe that ablating either token-level routing or sequence-level clustering leads to a consistent performance degradation relative to the full dual-stage configuration. For example, MBPP accuracy decreases from 23.55 to 21.72 without clustering and to 20.62 without token routing. The degradation in the sequence-level-only routing setup stems from the lack of token-level granularity, as all inputs are routed to a single expert based solely on coarse-grained features. In contrast, the token-only setup lacks global contextual signals, limiting the model’s ability to capture sequence-level semantics and leading to suboptimal expert assignment. These findings highlight the necessity of a routing hierarchy that combines sequence- and token-level mechanisms to enable effective expert specialization.

\section{Conclusion}
In this study, we introduce the Mixture-of-Clustered-Experts (MoCE), which incorporates a dual-stage routing mechanism that effectively leverages both sequence-level and token-level features. By incorporating a sequence-level expert group allocation and a token-level expert allocation, MoCE effectively manages inputs while maintaining computational efficiency. Our experimental results consistently show that MoCE outperforms baseline models, highlighting its ability to promote expert specialization and enhance generalization capability. Comprehensive evaluations further validated the effectiveness of our approach, establishing MoCE as a promising framework for managing complex inputs by encouraging expert specialization.

\section*{Limitations}
While MoCE introduces a novel dual-stage routing mechanism that improves model performance across various tasks, several limitations remain. First, due to computational resource constraints, we applied MoCE to adapter-based models to maintain efficiency. Applying this method directly to the full feed-forward network (FFN) layers for instruction tuning is left as future work. 

Second, interpretability is an ongoing challenge in MoE-based architectures. Although we employed k-means clustering, this method still does not address the explainability issues inherent in MoE systems. Even the interpretability of k-means clustering results is limited to a subset of distinctive clusters, such as mathematics and code.

Lastly, while MoCE has demonstrated strong results across tasks in mathematics, coding, and general knowledge, expanding the framework to more languages and specialized domains remains an open area for future work. Investigating its effectiveness in multilingual and domain-specific settings will be essential for broadening the applicability of MoCE.

\section*{Acknowledgments}
This work was supported by Institute for Information \& communications Technology Promotion(IITP) grant funded by the Korea government(MSIT) (RS-2024-00398115, Research on the reliability and coherence of outcomes produced by Generative AI) and this research was supported by Basic Science Research Program through the National Research Foundation of Korea(NRF) funded by the Ministry of Education(NRF-2021R1A6A1A03045425) and this work was supported by the Commercialization Promotion Agency for R\&D Outcomes(COMPA) grant funded by the Korea government(Ministry of Science and ICT)(2710086166).

\bibliography{custom}

\begin{thebibliography}{51}
\providecommand{\natexlab}[1]{#1}

\bibitem[{Amini et~al.(2019)Amini, Gabriel, Lin, Koncel-Kedziorski, Choi, and Hajishirzi}]{amini-etal-2019-mathqa}
Aida Amini, Saadia Gabriel, Shanchuan Lin, Rik Koncel-Kedziorski, Yejin Choi, and Hannaneh Hajishirzi. 2019.
\newblock \href {https://doi.org/10.18653/v1/N19-1245} {{M}ath{QA}: Towards interpretable math word problem solving with operation-based formalisms}.
\newblock In \emph{Proceedings of the 2019 Conference of the North {A}merican Chapter of the Association for Computational Linguistics: Human Language Technologies, Volume 1 (Long and Short Papers)}, pages 2357--2367, Minneapolis, Minnesota. Association for Computational Linguistics.

\bibitem[{Austin et~al.(2021)Austin, Odena, Nye, Bosma, Michalewski, Dohan, Jiang, Cai, Terry, Le et~al.}]{austin2021program}
Jacob Austin, Augustus Odena, Maxwell Nye, Maarten Bosma, Henryk Michalewski, David Dohan, Ellen Jiang, Carrie Cai, Michael Terry, Quoc Le, et~al. 2021.
\newblock Program synthesis with large language models.
\newblock \emph{arXiv preprint arXiv:2108.07732}.

\bibitem[{BehnamGhader et~al.(2024)BehnamGhader, Adlakha, Mosbach, Bahdanau, Chapados, and Reddy}]{behnamghader2024llmvec}
Parishad BehnamGhader, Vaibhav Adlakha, Marius Mosbach, Dzmitry Bahdanau, Nicolas Chapados, and Siva Reddy. 2024.
\newblock \href {https://openreview.net/forum?id=IW1PR7vEBf} {{LLM}2vec: Large language models are secretly powerful text encoders}.
\newblock In \emph{First Conference on Language Modeling}.

\bibitem[{Ben~Allal et~al.(2022)Ben~Allal, Muennighoff, Kumar~Umapathi, Lipkin, and von Werra}]{bigcode-evaluation-harness}
Loubna Ben~Allal, Niklas Muennighoff, Logesh Kumar~Umapathi, Ben Lipkin, and Leandro von Werra. 2022.
\newblock A framework for the evaluation of code generation models.
\newblock \url{https://github.com/bigcode-project/bigcode-evaluation-harness}.

\bibitem[{Cai et~al.(2024)Cai, Jiang, Wang, Tang, Kim, and Huang}]{cai2024survey}
Weilin Cai, Juyong Jiang, Fan Wang, Jing Tang, Sunghun Kim, and Jiayi Huang. 2024.
\newblock A survey on mixture of experts.
\newblock \emph{arXiv preprint arXiv:2407.06204}.

\bibitem[{Chen et~al.(2021)Chen, Tworek, Jun, Yuan, Pinto, Kaplan, Edwards, Burda, Joseph, Brockman et~al.}]{chen2021evaluating}
Mark Chen, Jerry Tworek, Heewoo Jun, Qiming Yuan, Henrique Ponde De~Oliveira Pinto, Jared Kaplan, Harri Edwards, Yuri Burda, Nicholas Joseph, Greg Brockman, et~al. 2021.
\newblock Evaluating large language models trained on code.
\newblock \emph{arXiv preprint arXiv:2107.03374}.

\bibitem[{Chen et~al.(2023)Chen, Zhang, JAISWAL, Liu, and Wang}]{chen2023sparse}
Tianlong Chen, Zhenyu Zhang, AJAY~KUMAR JAISWAL, Shiwei Liu, and Zhangyang Wang. 2023.
\newblock \href {https://openreview.net/forum?id=w1hwFUb_81} {Sparse moe as the new dropout: Scaling dense and self-slimmable transformers}.
\newblock In \emph{The Eleventh International Conference on Learning Representations}.

\bibitem[{Clark et~al.(2018)Clark, Cowhey, Etzioni, Khot, Sabharwal, Schoenick, and Tafjord}]{clark2018think}
Peter Clark, Isaac Cowhey, Oren Etzioni, Tushar Khot, Ashish Sabharwal, Carissa Schoenick, and Oyvind Tafjord. 2018.
\newblock Think you have solved question answering? try arc, the ai2 reasoning challenge.
\newblock \emph{arXiv preprint arXiv:1803.05457}.

\bibitem[{Cobbe et~al.(2021)Cobbe, Kosaraju, Bavarian, Chen, Jun, Kaiser, Plappert, Tworek, Hilton, Nakano et~al.}]{cobbe2021training}
Karl Cobbe, Vineet Kosaraju, Mohammad Bavarian, Mark Chen, Heewoo Jun, Lukasz Kaiser, Matthias Plappert, Jerry Tworek, Jacob Hilton, Reiichiro Nakano, et~al. 2021.
\newblock Training verifiers to solve math word problems.
\newblock \emph{arXiv preprint arXiv:2110.14168}.

\bibitem[{Dai et~al.(2024)Dai, Deng, Zhao, Xu, Gao, Chen, Li, Zeng, Yu, Wu, Xie, Li, Huang, Luo, Ruan, Sui, and Liang}]{dai-etal-2024-deepseekmoe}
Damai Dai, Chengqi Deng, Chenggang Zhao, R.x. Xu, Huazuo Gao, Deli Chen, Jiashi Li, Wangding Zeng, Xingkai Yu, Y.~Wu, Zhenda Xie, Y.k. Li, Panpan Huang, Fuli Luo, Chong Ruan, Zhifang Sui, and Wenfeng Liang. 2024.
\newblock \href {https://doi.org/10.18653/v1/2024.acl-long.70} {{D}eep{S}eek{M}o{E}: Towards ultimate expert specialization in mixture-of-experts language models}.
\newblock In \emph{Proceedings of the 62nd Annual Meeting of the Association for Computational Linguistics (Volume 1: Long Papers)}, pages 1280--1297, Bangkok, Thailand. Association for Computational Linguistics.

\bibitem[{Dai et~al.(2022)Dai, Tang, Liu, Tan, Zhou, Wang, Feng, Zhang, Hu, and Shi}]{dai2022one}
Yong Dai, Duyu Tang, Liangxin Liu, Minghuan Tan, Cong Zhou, Jingquan Wang, Zhangyin Feng, Fan Zhang, Xueyu Hu, and Shuming Shi. 2022.
\newblock One model, multiple modalities: A sparsely activated approach for text, sound, image, video and code.
\newblock \emph{arXiv preprint arXiv:2205.06126}.

\bibitem[{Diao et~al.(2023)Diao, Xu, Xu, Wang, and Zhang}]{diao-etal-2023-mixture}
Shizhe Diao, Tianyang Xu, Ruijia Xu, Jiawei Wang, and Tong Zhang. 2023.
\newblock \href {https://doi.org/10.18653/v1/2023.acl-long.280} {Mixture-of-domain-adapters: Decoupling and injecting domain knowledge to pre-trained language models{'} memories}.
\newblock In \emph{Proceedings of the 61st Annual Meeting of the Association for Computational Linguistics (Volume 1: Long Papers)}, pages 5113--5129, Toronto, Canada. Association for Computational Linguistics.

\bibitem[{Dou et~al.(2024)Dou, Zhou, Liu, Gao, Shen, Xiong, Zhou, Wang, Xi, Fan et~al.}]{dou2024loramoe}
Shihan Dou, Enyu Zhou, Yan Liu, Songyang Gao, Wei Shen, Limao Xiong, Yuhao Zhou, Xiao Wang, Zhiheng Xi, Xiaoran Fan, et~al. 2024.
\newblock Loramoe: Alleviating world knowledge forgetting in large language models via moe-style plugin.
\newblock In \emph{Proceedings of the 62nd Annual Meeting of the Association for Computational Linguistics (Volume 1: Long Papers)}, pages 1932--1945.

\bibitem[{Fan et~al.(2024)Fan, Messmer, and Jaggi}]{fan2024towards}
Dongyang Fan, Bettina Messmer, and Martin Jaggi. 2024.
\newblock \href {https://openreview.net/forum?id=ebPKyb6r9F} {{TOWARDS} {AN} {EMPIRICAL} {UNDERSTANDING} {OF} {MOE} {DESIGN} {CHOICES}}.
\newblock In \emph{ICLR 2024 Workshop on Mathematical and Empirical Understanding of Foundation Models}.

\bibitem[{Gao et~al.(2024)Gao, Tow, Abbasi, Biderman, Black, DiPofi, Foster, Golding, Hsu, Le~Noac'h, Li, McDonell, Muennighoff, Ociepa, Phang, Reynolds, Schoelkopf, Skowron, Sutawika, Tang, Thite, Wang, Wang, and Zou}]{eval-harness}
Leo Gao, Jonathan Tow, Baber Abbasi, Stella Biderman, Sid Black, Anthony DiPofi, Charles Foster, Laurence Golding, Jeffrey Hsu, Alain Le~Noac'h, Haonan Li, Kyle McDonell, Niklas Muennighoff, Chris Ociepa, Jason Phang, Laria Reynolds, Hailey Schoelkopf, Aviya Skowron, Lintang Sutawika, Eric Tang, Anish Thite, Ben Wang, Kevin Wang, and Andy Zou. 2024.
\newblock \href {https://doi.org/10.5281/zenodo.12608602} {A framework for few-shot language model evaluation}.

\bibitem[{Gou et~al.(2023)Gou, Liu, Chen, Hong, Xu, Li, Yeung, Kwok, and Zhang}]{gou2023mixture}
Yunhao Gou, Zhili Liu, Kai Chen, Lanqing Hong, Hang Xu, Aoxue Li, Dit-Yan Yeung, James~T Kwok, and Yu~Zhang. 2023.
\newblock Mixture of cluster-conditional lora experts for vision-language instruction tuning.
\newblock \emph{arXiv preprint arXiv:2312.12379}.

\bibitem[{Gururangan et~al.(2022)Gururangan, Lewis, Holtzman, Smith, and Zettlemoyer}]{gururangan-etal-2022-demix}
Suchin Gururangan, Mike Lewis, Ari Holtzman, Noah~A. Smith, and Luke Zettlemoyer. 2022.
\newblock \href {https://doi.org/10.18653/v1/2022.naacl-main.407} {{DEM}ix layers: Disentangling domains for modular language modeling}.
\newblock In \emph{Proceedings of the 2022 Conference of the North American Chapter of the Association for Computational Linguistics: Human Language Technologies}, pages 5557--5576, Seattle, United States. Association for Computational Linguistics.

\bibitem[{Houlsby et~al.(2019)Houlsby, Giurgiu, Jastrzebski, Morrone, De~Laroussilhe, Gesmundo, Attariyan, and Gelly}]{houlsby2019parameter}
Neil Houlsby, Andrei Giurgiu, Stanislaw Jastrzebski, Bruna Morrone, Quentin De~Laroussilhe, Andrea Gesmundo, Mona Attariyan, and Sylvain Gelly. 2019.
\newblock Parameter-efficient transfer learning for nlp.
\newblock In \emph{International conference on machine learning}, pages 2790--2799. PMLR.

\bibitem[{Jiang et~al.(2024)Jiang, Sablayrolles, Roux, Mensch, Savary, Bamford, Chaplot, Casas, Hanna, Bressand et~al.}]{jiang2024mixtral}
Albert~Q Jiang, Alexandre Sablayrolles, Antoine Roux, Arthur Mensch, Blanche Savary, Chris Bamford, Devendra~Singh Chaplot, Diego de~las Casas, Emma~Bou Hanna, Florian Bressand, et~al. 2024.
\newblock Mixtral of experts.
\newblock \emph{arXiv preprint arXiv:2401.04088}.

\bibitem[{Komatsuzaki et~al.(2023)Komatsuzaki, Puigcerver, Lee-Thorp, Ruiz, Mustafa, Ainslie, Tay, Dehghani, and Houlsby}]{komatsuzaki2023sparse}
Aran Komatsuzaki, Joan Puigcerver, James Lee-Thorp, Carlos~Riquelme Ruiz, Basil Mustafa, Joshua Ainslie, Yi~Tay, Mostafa Dehghani, and Neil Houlsby. 2023.
\newblock \href {https://openreview.net/forum?id=T5nUQDrM4u} {Sparse upcycling: Training mixture-of-experts from dense checkpoints}.
\newblock In \emph{The Eleventh International Conference on Learning Representations}.

\bibitem[{Kudugunta et~al.(2021)Kudugunta, Huang, Bapna, Krikun, Lepikhin, Luong, and Firat}]{kudugunta-etal-2021-beyond-distillation}
Sneha Kudugunta, Yanping Huang, Ankur Bapna, Maxim Krikun, Dmitry Lepikhin, Minh-Thang Luong, and Orhan Firat. 2021.
\newblock \href {https://doi.org/10.18653/v1/2021.findings-emnlp.304} {Beyond distillation: Task-level mixture-of-experts for efficient inference}.
\newblock In \emph{Findings of the Association for Computational Linguistics: EMNLP 2021}, pages 3577--3599, Punta Cana, Dominican Republic. Association for Computational Linguistics.

\bibitem[{Lian et~al.(2023)Lian, Wang, Goodson, Pentland, Cook, and Vong}]{lian2023slimorca}
Wing Lian, Guan Wang, Bleys Goodson, Eugene Pentland, Austin Cook, and Chanvichet Vong. 2023.
\newblock Slimorca: An open dataset of gpt-4 augmented flan reasoning traces, with verification.
\newblock \emph{Slimorca: An open dataset of gpt-4 augmented flan reasoning traces, with verification}, 5.

\bibitem[{Longpre et~al.(2023)Longpre, Hou, Vu, Webson, Chung, Tay, Zhou, Le, Zoph, Wei, and Roberts}]{pmlr-v202-longpre23a}
Shayne Longpre, Le~Hou, Tu~Vu, Albert Webson, Hyung~Won Chung, Yi~Tay, Denny Zhou, Quoc~V Le, Barret Zoph, Jason Wei, and Adam Roberts. 2023.
\newblock \href {https://proceedings.mlr.press/v202/longpre23a.html} {The flan collection: Designing data and methods for effective instruction tuning}.
\newblock In \emph{Proceedings of the 40th International Conference on Machine Learning}, volume 202 of \emph{Proceedings of Machine Learning Research}, pages 22631--22648. PMLR.

\bibitem[{Ma et~al.(2018)Ma, Zhao, Yi, Chen, Hong, and Chi}]{ma2018modeling}
Jiaqi Ma, Zhe Zhao, Xinyang Yi, Jilin Chen, Lichan Hong, and Ed~H Chi. 2018.
\newblock Modeling task relationships in multi-task learning with multi-gate mixture-of-experts.
\newblock In \emph{Proceedings of the 24th ACM SIGKDD international conference on knowledge discovery \& data mining}, pages 1930--1939.

\bibitem[{Muqeeth et~al.(2024)Muqeeth, Liu, and Raffel}]{muqeeth2024soft}
Mohammed Muqeeth, Haokun Liu, and Colin Raffel. 2024.
\newblock \href {https://openreview.net/forum?id=QHzzAU7Qf9} {Soft merging of experts with adaptive routing}.

\bibitem[{Ostapenko et~al.(2023)Ostapenko, Caccia, Su, Le~Roux, Charlin, and Sordoni}]{ostapenko2023case}
Oleksiy Ostapenko, Lucas Caccia, Zhan Su, Nicolas Le~Roux, Laurent Charlin, and Alessandro Sordoni. 2023.
\newblock A case study of instruction tuning with mixture of parameter-efficient experts.
\newblock In \emph{NeurIPS 2023 Workshop on Instruction Tuning and Instruction Following}.

\bibitem[{Peng et~al.(2023)Peng, Li, He, Galley, and Gao}]{peng2023instruction}
Baolin Peng, Chunyuan Li, Pengcheng He, Michel Galley, and Jianfeng Gao. 2023.
\newblock Instruction tuning with gpt-4.
\newblock \emph{arXiv preprint arXiv:2304.03277}.

\bibitem[{Puigcerver et~al.(2024)Puigcerver, Ruiz, Mustafa, and Houlsby}]{puigcerver2024from}
Joan Puigcerver, Carlos~Riquelme Ruiz, Basil Mustafa, and Neil Houlsby. 2024.
\newblock \href {https://openreview.net/forum?id=jxpsAj7ltE} {From sparse to soft mixtures of experts}.
\newblock In \emph{The Twelfth International Conference on Learning Representations}.

\bibitem[{Raposo et~al.(2024)Raposo, Ritter, Richards, Lillicrap, Humphreys, and Santoro}]{raposo2024mixture}
David Raposo, Sam Ritter, Blake Richards, Timothy Lillicrap, Peter~Conway Humphreys, and Adam Santoro. 2024.
\newblock Mixture-of-depths: Dynamically allocating compute in transformer-based language models.
\newblock \emph{arXiv preprint arXiv:2404.02258}.

\bibitem[{Ren et~al.(2023)Ren, Zhou, Meng, Huang, Wang, Wang, Li, Zhang, Podolskiy, Arshinov et~al.}]{ren2023pangu}
Xiaozhe Ren, Pingyi Zhou, Xinfan Meng, Xinjing Huang, Yadao Wang, Weichao Wang, Pengfei Li, Xiaoda Zhang, Alexander Podolskiy, Grigory Arshinov, et~al. 2023.
\newblock Pangu-$\{$$\backslash$Sigma$\}$: Towards trillion parameter language model with sparse heterogeneous computing.
\newblock \emph{arXiv preprint arXiv:2303.10845}.

\bibitem[{Sakaguchi et~al.(2021)Sakaguchi, Bras, Bhagavatula, and Choi}]{10.1145/3474381}
Keisuke Sakaguchi, Ronan~Le Bras, Chandra Bhagavatula, and Yejin Choi. 2021.
\newblock \href {https://doi.org/10.1145/3474381} {Winogrande: an adversarial winograd schema challenge at scale}.
\newblock \emph{Commun. ACM}, 64(9):99–106.

\bibitem[{Sarkar et~al.(2024)Sarkar, Lausen, Cevher, Zha, Brox, and Karypis}]{sarkar2024revisiting}
Soumajyoti Sarkar, Leonard Lausen, Volkan Cevher, Sheng Zha, Thomas Brox, and George Karypis. 2024.
\newblock Revisiting smoe language models by evaluating inefficiencies with task specific expert pruning.
\newblock \emph{arXiv preprint arXiv:2409.01483}.

\bibitem[{Shazeer et~al.(2017)Shazeer, Mirhoseini, Maziarz, Davis, Le, Hinton, and Dean}]{shazeer2017}
Noam Shazeer, *Azalia Mirhoseini, *Krzysztof Maziarz, Andy Davis, Quoc Le, Geoffrey Hinton, and Jeff Dean. 2017.
\newblock \href {https://openreview.net/forum?id=B1ckMDqlg} {Outrageously large neural networks: The sparsely-gated mixture-of-experts layer}.
\newblock In \emph{International Conference on Learning Representations}.

\bibitem[{Shen et~al.(2024)Shen, Hou, Zhou, Du, Longpre, Wei, Chung, Zoph, Fedus, Chen, Vu, Wu, Chen, Webson, Li, Zhao, Yu, Keutzer, Darrell, and Zhou}]{shen2024mixtureofexperts}
Sheng Shen, Le~Hou, Yanqi Zhou, Nan Du, Shayne Longpre, Jason Wei, Hyung~Won Chung, Barret Zoph, William Fedus, Xinyun Chen, Tu~Vu, Yuexin Wu, Wuyang Chen, Albert Webson, Yunxuan Li, Vincent~Y Zhao, Hongkun Yu, Kurt Keutzer, Trevor Darrell, and Denny Zhou. 2024.
\newblock \href {https://openreview.net/forum?id=6mLjDwYte5} {Mixture-of-experts meets instruction tuning: A winning combination for large language models}.
\newblock In \emph{The Twelfth International Conference on Learning Representations}.

\bibitem[{Su et~al.(2023)Su, Shi, Kasai, Wang, Hu, Ostendorf, Yih, Smith, Zettlemoyer, and Yu}]{su-etal-2023-one}
Hongjin Su, Weijia Shi, Jungo Kasai, Yizhong Wang, Yushi Hu, Mari Ostendorf, Wen-tau Yih, Noah~A. Smith, Luke Zettlemoyer, and Tao Yu. 2023.
\newblock \href {https://doi.org/10.18653/v1/2023.findings-acl.71} {One embedder, any task: Instruction-finetuned text embeddings}.
\newblock In \emph{Findings of the Association for Computational Linguistics: ACL 2023}, pages 1102--1121, Toronto, Canada. Association for Computational Linguistics.

\bibitem[{Sukhbaatar et~al.(2024)Sukhbaatar, Golovneva, Sharma, Xu, Lin, Roziere, Kahn, Li, tau Yih, Weston, and Li}]{sukhbaatar2024branchtrainmix}
Sainbayar Sukhbaatar, Olga Golovneva, Vasu Sharma, Hu~Xu, Xi~Victoria Lin, Baptiste Roziere, Jacob Kahn, Shang-Wen Li, Wen tau Yih, Jason~E Weston, and Xian Li. 2024.
\newblock \href {https://openreview.net/forum?id=nqLAuMOF6n} {Branch-train-mix: Mixing expert {LLM}s into a mixture-of-experts {LLM}}.
\newblock In \emph{First Conference on Language Modeling}.

\bibitem[{Suzgun et~al.(2022)Suzgun, Scales, Schärli, Gehrmann, Tay, Chung, Chowdhery, Le, Chi, Zhou, and Wei}]{suzgun2022challengingbigbenchtaskschainofthought}
Mirac Suzgun, Nathan Scales, Nathanael Schärli, Sebastian Gehrmann, Yi~Tay, Hyung~Won Chung, Aakanksha Chowdhery, Quoc~V. Le, Ed~H. Chi, Denny Zhou, and Jason Wei. 2022.
\newblock \href {https://arxiv.org/abs/2210.09261} {Challenging big-bench tasks and whether chain-of-thought can solve them}.
\newblock \emph{Preprint}, arXiv:2210.09261.

\bibitem[{Touvron et~al.(2023)Touvron, Martin, Stone, Albert, Almahairi, Babaei, Bashlykov, Batra, Bhargava, Bhosale et~al.}]{touvron2023llama}
Hugo Touvron, Louis Martin, Kevin Stone, Peter Albert, Amjad Almahairi, Yasmine Babaei, Nikolay Bashlykov, Soumya Batra, Prajjwal Bhargava, Shruti Bhosale, et~al. 2023.
\newblock Llama 2: Open foundation and fine-tuned chat models.
\newblock \emph{arXiv preprint arXiv:2307.09288}.

\bibitem[{Wang et~al.(2022{\natexlab{a}})Wang, Yang, Huang, Jiao, Yang, Jiang, Majumder, and Wei}]{wang2022text}
Liang Wang, Nan Yang, Xiaolong Huang, Binxing Jiao, Linjun Yang, Daxin Jiang, Rangan Majumder, and Furu Wei. 2022{\natexlab{a}}.
\newblock Text embeddings by weakly-supervised contrastive pre-training.
\newblock \emph{arXiv preprint arXiv:2212.03533}.

\bibitem[{Wang et~al.(2022{\natexlab{b}})Wang, Agarwal, Mukherjee, Liu, Gao, Awadallah, and Gao}]{wang-etal-2022-adamix}
Yaqing Wang, Sahaj Agarwal, Subhabrata Mukherjee, Xiaodong Liu, Jing Gao, Ahmed~Hassan Awadallah, and Jianfeng Gao. 2022{\natexlab{b}}.
\newblock \href {https://doi.org/10.18653/v1/2022.emnlp-main.388} {{A}da{M}ix: Mixture-of-adaptations for parameter-efficient model tuning}.
\newblock In \emph{Proceedings of the 2022 Conference on Empirical Methods in Natural Language Processing}, pages 5744--5760, Abu Dhabi, United Arab Emirates. Association for Computational Linguistics.

\bibitem[{Wang et~al.(2022{\natexlab{c}})Wang, Mishra, Alipoormolabashi, Kordi, Mirzaei, Naik, Ashok, Dhanasekaran, Arunkumar, Stap, Pathak, Karamanolakis, Lai, Purohit, Mondal, Anderson, Kuznia, Doshi, Pal, Patel, Moradshahi, Parmar, Purohit, Varshney, Kaza, Verma, Puri, Karia, Doshi, Sampat, Mishra, Reddy~A, Patro, Dixit, and Shen}]{wang-etal-2022-super}
Yizhong Wang, Swaroop Mishra, Pegah Alipoormolabashi, Yeganeh Kordi, Amirreza Mirzaei, Atharva Naik, Arjun Ashok, Arut~Selvan Dhanasekaran, Anjana Arunkumar, David Stap, Eshaan Pathak, Giannis Karamanolakis, Haizhi Lai, Ishan Purohit, Ishani Mondal, Jacob Anderson, Kirby Kuznia, Krima Doshi, Kuntal~Kumar Pal, Maitreya Patel, Mehrad Moradshahi, Mihir Parmar, Mirali Purohit, Neeraj Varshney, Phani~Rohitha Kaza, Pulkit Verma, Ravsehaj~Singh Puri, Rushang Karia, Savan Doshi, Shailaja~Keyur Sampat, Siddhartha Mishra, Sujan Reddy~A, Sumanta Patro, Tanay Dixit, and Xudong Shen. 2022{\natexlab{c}}.
\newblock \href {https://doi.org/10.18653/v1/2022.emnlp-main.340} {Super-{N}atural{I}nstructions: Generalization via declarative instructions on 1600+ {NLP} tasks}.
\newblock In \emph{Proceedings of the 2022 Conference on Empirical Methods in Natural Language Processing}, pages 5085--5109, Abu Dhabi, United Arab Emirates. Association for Computational Linguistics.

\bibitem[{Wang et~al.(2024)Wang, Ma, Zhang, Ni, Chandra, Guo, Ren, Arulraj, He, Jiang, Li, Ku, Wang, Zhuang, Fan, Yue, and Chen}]{wang2024mmluprorobustchallengingmultitask}
Yubo Wang, Xueguang Ma, Ge~Zhang, Yuansheng Ni, Abhranil Chandra, Shiguang Guo, Weiming Ren, Aaran Arulraj, Xuan He, Ziyan Jiang, Tianle Li, Max Ku, Kai Wang, Alex Zhuang, Rongqi Fan, Xiang Yue, and Wenhu Chen. 2024.
\newblock \href {https://arxiv.org/abs/2406.01574} {Mmlu-pro: A more robust and challenging multi-task language understanding benchmark}.
\newblock \emph{Preprint}, arXiv:2406.01574.

\bibitem[{Wei et~al.(2024)Wei, Wang, Liu, Ding, and Zhang}]{wei2024magicoder}
Yuxiang Wei, Zhe Wang, Jiawei Liu, Yifeng Ding, and Lingming Zhang. 2024.
\newblock Magicoder: Empowering code generation with oss-instruct.
\newblock In \emph{Forty-first International Conference on Machine Learning}.

\bibitem[{Wu et~al.(2024{\natexlab{a}})Wu, Zheng, and Yu}]{wu2024parameter}
Haoyuan Wu, Haisheng Zheng, and Bei Yu. 2024{\natexlab{a}}.
\newblock Parameter-efficient sparsity crafting from dense to mixture-of-experts for instruction tuning on general tasks.
\newblock \emph{arXiv preprint arXiv:2401.02731}.

\bibitem[{Wu et~al.(2024{\natexlab{b}})Wu, Huang, and Wei}]{wu2024mixture}
Xun Wu, Shaohan Huang, and Furu Wei. 2024{\natexlab{b}}.
\newblock \href {https://openreview.net/forum?id=uWvKBCYh4S} {Mixture of lo{RA} experts}.
\newblock In \emph{The Twelfth International Conference on Learning Representations}.

\bibitem[{Xue et~al.(2024)Xue, Zheng, Fu, Ni, Zheng, Zhou, and You}]{xue2024openmoe}
Fuzhao Xue, Zian Zheng, Yao Fu, Jinjie Ni, Zangwei Zheng, Wangchunshu Zhou, and Yang You. 2024.
\newblock Openmoe: An early effort on open mixture-of-experts language models.
\newblock \emph{arXiv preprint arXiv:2402.01739}.

\bibitem[{Zadouri et~al.(2024)Zadouri, {\"U}st{\"u}n, Ahmadian, Ermis, Locatelli, and Hooker}]{zadouri2024pushing}
Ted Zadouri, Ahmet {\"U}st{\"u}n, Arash Ahmadian, Beyza Ermis, Acyr Locatelli, and Sara Hooker. 2024.
\newblock \href {https://openreview.net/forum?id=EvDeiLv7qc} {Pushing mixture of experts to the limit: Extremely parameter efficient moe for instruction tuning}.
\newblock In \emph{The Twelfth International Conference on Learning Representations}.

\bibitem[{Zhao et~al.(2023)Zhao, Chen, Cheng, and Chen}]{zhao2023sparse}
Xinyu Zhao, Xuxi Chen, Yu~Cheng, and Tianlong Chen. 2023.
\newblock Sparse moe with language guided routing for multilingual machine translation.
\newblock In \emph{The Twelfth International Conference on Learning Representations}.

\bibitem[{Zhong et~al.(2022)Zhong, Chi, Gu, Wang, Yu, and Tang}]{zhong2022meta}
Tao Zhong, Zhixiang Chi, Li~Gu, Yang Wang, Yuanhao Yu, and Jin Tang. 2022.
\newblock Meta-dmoe: Adapting to domain shift by meta-distillation from mixture-of-experts.
\newblock \emph{Advances in Neural Information Processing Systems}, 35:22243--22257.

\bibitem[{Zhu et~al.(2024)Zhu, Dong, Qu, Ruan, Chen, and Cheng}]{zhu2024dynamic}
Tong Zhu, Daize Dong, Xiaoye Qu, Jiacheng Ruan, Wenliang Chen, and Yu~Cheng. 2024.
\newblock Dynamic data mixing maximizes instruction tuning for mixture-of-experts.
\newblock \emph{arXiv preprint arXiv:2406.11256}.

\bibitem[{Zoph et~al.(2022)Zoph, Bello, Kumar, Du, Huang, Dean, Shazeer, and Fedus}]{zoph2022st}
Barret Zoph, Irwan Bello, Sameer Kumar, Nan Du, Yanping Huang, Jeff Dean, Noam Shazeer, and William Fedus. 2022.
\newblock St-moe: Designing stable and transferable sparse expert models.
\newblock \emph{arXiv preprint arXiv:2202.08906}.

\end{thebibliography}

\appendix
\section{Implementation Details}\label{appendix:implementation_details}
The hyperparameters used in our experiments are as follows: the adapter dimension is set to 64, with an MoE scaling factor of 1, and the number of experts is set to 4 per cluster. The maximum sequence length is 512, and we apply top-2 expert selection for expert routing. The learning rate is set to $2\times10^{-4}$, and a batch size of 32 is used. All models are trained for 1 epoch on eight 80G A100 GPUs. 

Two embedding models are used for clustering: E5 (\texttt{intfloat/e5-large-v2}) and Instructor (\texttt{hkunlp/instructor-xl}), both initialized from Huggingface checkpoints. Similarly, three models are used for instruction tuning: LLaMA2 (\texttt{meta-llama/Llama-2-7b-chat-hf}), Mistral (\texttt{mistralai/Mistral-7B-Instruct-v0}), and Qwen (\texttt{Qwen/Qwen2-7B-Instruct}), also initialized from Huggingface checkpoints. 

\section{Details of the Elbow Method} \label{app:elbow_method}
We employ the elbow method to minimize manual configuration and mitigate overfitting when clustering instruction tuning datasets with high variance. The optimal number of clusters k is determined through the following steps:
\begin{itemize}
\item Step 1: We iteratively train k-means clustering models, gradually increasing the number of clusters. In our experiment, the number of clusters is incremented from 1 to 10.
\item Step 2: We measure and record the within-cluster sum of squared errors (SSE) for each cluster count. As the number of clusters increases, SSE progressively decreases.
\item Step 3: We determine the point at which the decrease in SSE clearly begins to plateau. This indicates the elbow point, which is an optimal number of clusters.
\end{itemize}

\section{Embedding Results}\label{app:embed}
We present a visualization of the clustering results obtained using both E5 and Instructor embeddings, with dimensionality reduction performed via t-SNE. The resulting visualizations are shown in Figure~\ref{fig:embed}. In both cases, the embeddings exhibit strong convergence around well-defined centroids. The clear separation among clusters suggests that the model effectively captures latent semantic distinctions across input types, facilitating more targeted expert activation. These results provide qualitative support for the expert specialization mechanism in MoCE.

\end{document}